\documentclass[conference]{IEEEtran}
\IEEEoverridecommandlockouts
\usepackage{cite}
\usepackage{amsmath,amssymb,amsfonts}
\usepackage{algorithmic}
\usepackage{graphicx}
\usepackage{textcomp}
\usepackage{xcolor}
\usepackage{makecell}
\usepackage[hidelinks]{hyperref}
\usepackage{bbding}
\usepackage{multirow}
\usepackage{booktabs}
\usepackage{footmisc}
\usepackage{amsfonts,amssymb,amsmath}
\def\BibTeX{{\rm B\kern-.05em{\sc i\kern-.025em b}\kern-.08em
    T\kern-.1667em\lower.7ex\hbox{E}\kern-.125emX}}
\begin{document}

\title{Multi-modality Anomaly Segmentation on the Road}

\author{\IEEEauthorblockN{Heng Gao$^1$, Zhuolin He$^2$, Shoumeng Qiu$^2$, Xiangyang Xue$^{2, \ast}$\thanks{*Corresponding author.}, Jian Pu$^1$}
\IEEEauthorblockA{$^1$\textit{Institute of Science and Technology for Brain-inspired Intelligence, Fudan University, Shanghai, China} \\
\textit{$^2$School of Computer Science, Fudan University, Shanghai, China}\\
\{hgao22,zlhe22\}@m.fudan.edu.cn, skyshoumeng@163.com, \{xyxue, jianpu\}@fudan.edu.cn}
}

\maketitle

\begin{abstract}
Semantic segmentation allows autonomous driving cars to understand the surroundings of the vehicle comprehensively. However, it is also crucial for the model to detect obstacles that may jeopardize the safety of autonomous driving systems. Based on our experiments, we find that current uni-modal anomaly segmentation frameworks tend to produce high anomaly scores for non-anomalous regions in images. Motivated by this empirical finding, we develop a multi-modal uncertainty-based anomaly segmentation framework, named MMRAS+, for autonomous driving systems. MMRAS+ effectively reduces the high anomaly outputs of non-anomalous classes by introducing text-modal using the CLIP text encoder. Indeed, MMRAS+ is the first multi-modal anomaly segmentation solution for autonomous driving. Moreover, we develop an ensemble module to further boost the anomaly segmentation performance. Experiments on RoadAnomaly, SMIYC, and Fishyscapes validation datasets demonstrate the superior performance of our method. The code is available in \url{https://github.com/HengGao12/MMRAS_plus}.
\end{abstract}

\begin{IEEEkeywords}
Deep learning, anomaly segmentation, multi-modal, autonomous driving
\end{IEEEkeywords}

\section{Introduction}
\label{sec:intro}

Semantic segmentation \cite{Long_2015_CVPR, zhao2017pyramid, chen2018encoder, xie2021segformer, zheng2021rethinking, cheng2021per, Chen_2023_CVPR} is a powerful tool for autonomous driving systems to understand the surrounding environments of the vehicle. However, when deployed in the real world, autonomous driving models may encounter unknown objects that they have not seen during training. This problem could pose risks to the safety of autonomous driving cars. Therefore, in recent years, researchers have proposed many anomaly segmentation algorithms \cite{ackermann2023maskomaly, basart2022scaling, jung2021standardized, rai2023unmasking, Di_Biase_2021_CVPR} to help the model segment anomalies on the road.

\begin{figure}[t]
\centering
\includegraphics[width=\linewidth]{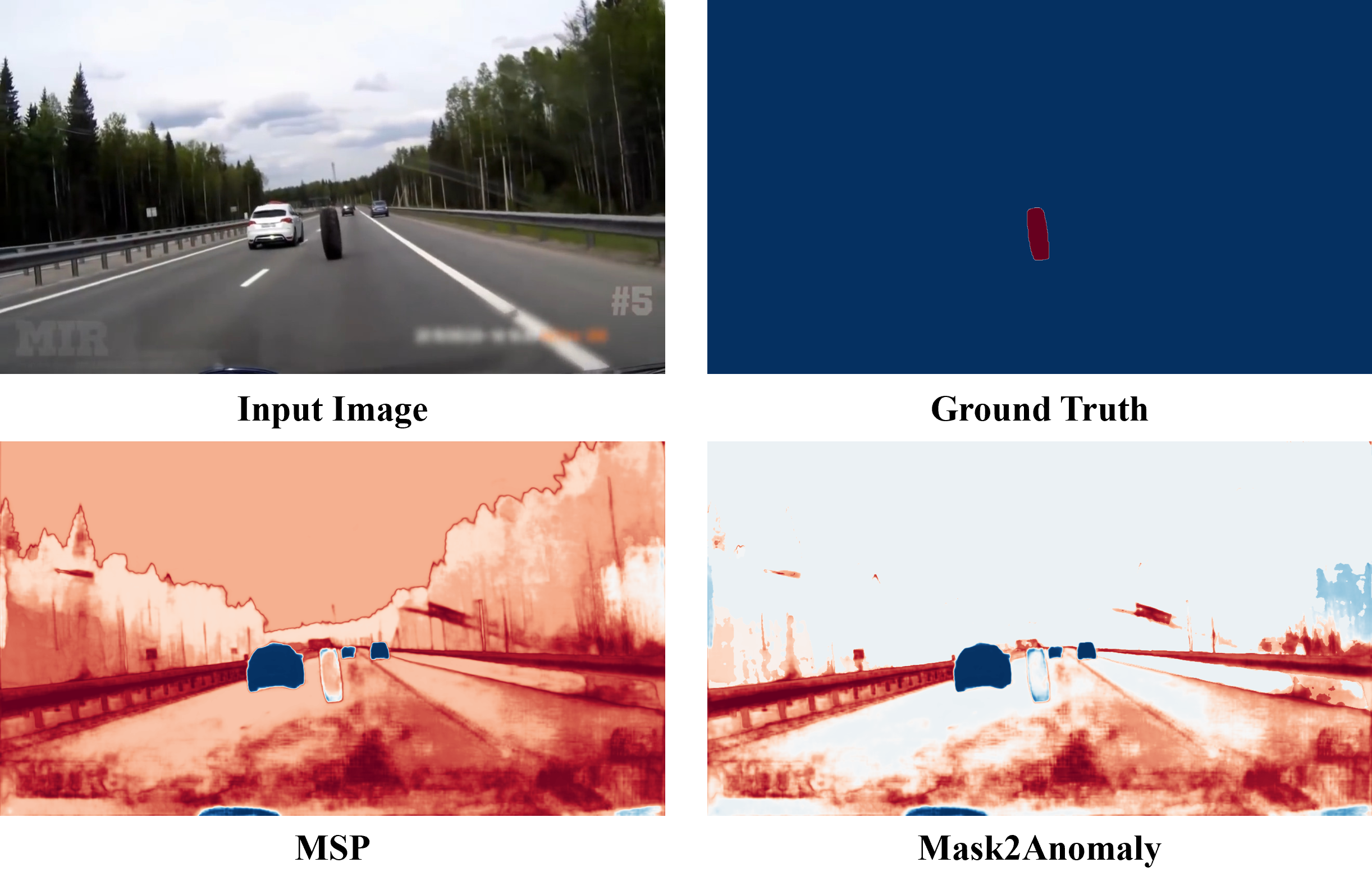}
\caption{The limitation of current uni-modal based anomaly segmentation approaches for autonomous driving. The uni-modal-based methods may tend to produce high anomaly predictions for non-anomalous regions in the image. See the second row in the above figure, MSP \cite{hendrycks2017a} and Mask2Anomaly \cite{rai2023unmasking} output high anomaly scores for the sky and trees classes in the image on RoadAnomaly \cite{lis2019detecting} validation dataset. Note that in the Ground Truth image, the red parts represent anomalous objects and the blue parts represent non-anomalous regions. Besides, \textbf{the closer the color is to red, the more likely it is to be an anomaly, the closer the color is to blue, the less likely it is to be an anomaly}.}
\label{fig:motivation}
\end{figure}

Nevertheless, based on our experiments, we observe that the current uni-modal based anomaly segmentation methods \cite{hendrycks2017a,liang2018enhancing,rai2023unmasking} tend to produce high anomaly scores for non-anomalous regions in autonomous driving scenes. For instance, as shown in the second row of Fig.~\ref{fig:motivation}, using the architecture proposed in \cite{rai2023unmasking}, both MSP and Mask2Anomaly output high anomaly scores for the tree and sky classes, which belong to non-anomalous objects.

Therefore, from the perspective of multi-modal learning, we propose introducing text-modal into the anomaly segmentation framework to mitigate this issue. The intuition here is to leverage the information from in-distribution classes to \textquotedblleft tell\textquotedblright~ the model in which objects in the image are within the training distribution. Besides, the main challenge in achieving our goal is to find a feasible way to incorporate the text branch into the current image-based anomaly segmentation network. To address this challenge, we present a \textbf{M}ulti-\textbf{M}odal uncertainty-based \textbf{R}oad \textbf{A}nomaly \textbf{S}egmentation framework, named MMRAS+, for autonomous driving scenes. In MMRAS+, we first use a CLIP  \cite{radford2021learning} text encoder to extract text embeddings from text prompts. Afterward, we take these embeddings as the weight of a segmentation head to obtain the text-image logits. Then, we perform a weighted summation of the text-image logits and the outputs of Mask2Anomaly to get the final logits. We use the final logits to compute the anomaly score maps. To the best of our knowledge, we present the first multi-modal anomaly segmentation solution for autonomous driving scenes.

Moreover, to further enhance the anomaly segmentation performance, we develop an ensemble module that also computes a weighted sum of several different anomaly score outputs. The idea of this ensemble module stems from a classic machine learning algorithm called AdaBoost \cite{freund1997decision}, which ensembles many weakly learnable algorithms to attain a strongly learnable algorithm. Herein, similarly, we create a strong anomaly detector by integrating different \textquotedblleft weak\textquotedblright~ anomaly scores. The quantitative and qualitative experiments on RoadAnomaly, SegmentMeIfYouCan (SMIYC) and Fishyscapes validation datasets demonstrate that our multi-modal anomaly segmentation framework effectively reduces the anomaly scores of non-anomalous regions in images of autonomous driving scenes.

In brief, the main contributions of our paper are:
\begin{itemize}
    \item We propose a multi-modal uncertainty-based anomaly segmentation approach named MMRAS+, which mitigates the issue of producing high anomaly scores for non-anomalous regions in autonomous driving scenes by introducing text-modal into the uni-modal anomaly segmentation framework.
    \item We investigate the application of MaskCLIP \cite{zhou2022extract} in segmenting anomalies for road scenes. In addition, we develop an ensemble module to further boost the anomaly segmentation performance of our framework.
    \item The quantitative and qualitative experiments on RoadAnomaly, SMIYC, and Fishyscapes validation datasets demonstrate the effectiveness of our method.
\end{itemize}

\begin{figure*}[t]
\centering
\includegraphics[width=\linewidth]{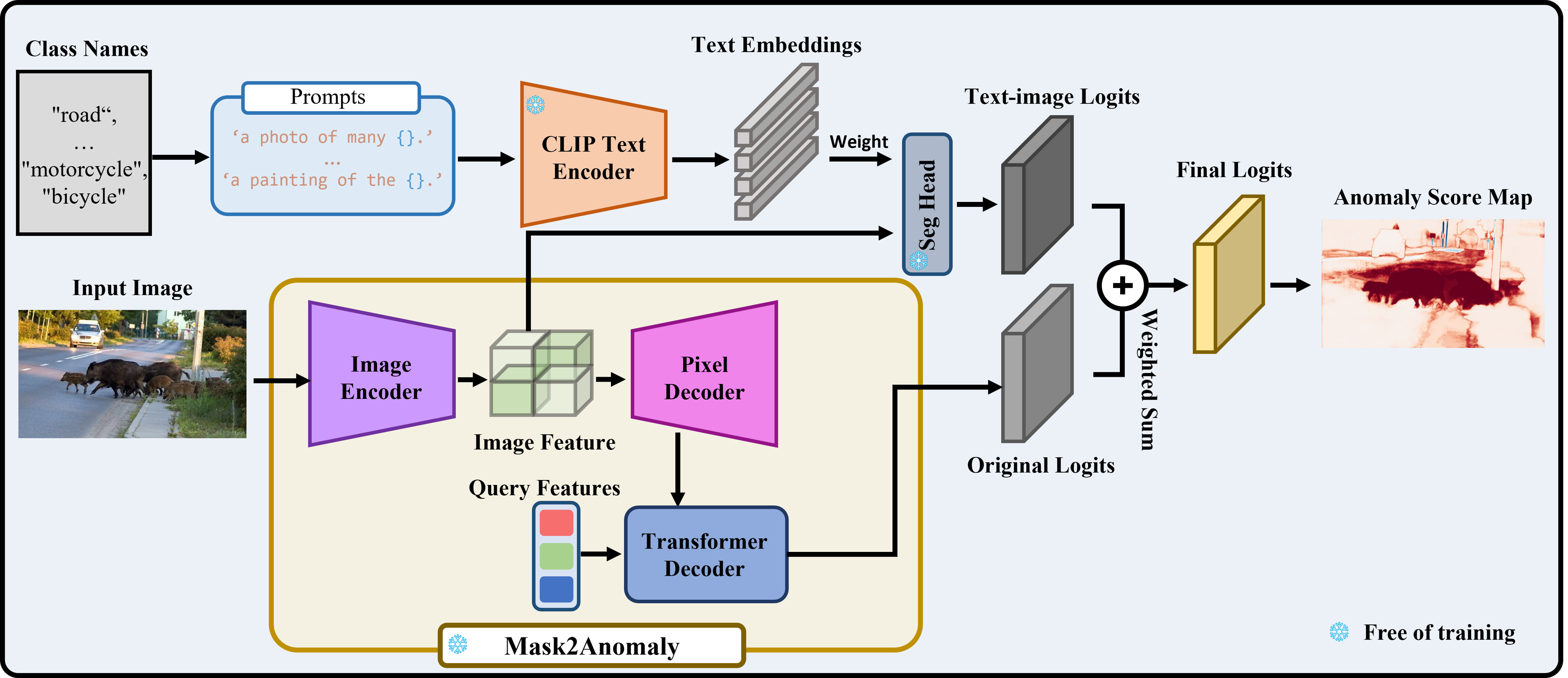}
\caption{The overview of MMRAS for anomaly segmentation in autonomous driving scenes. MMRAS first uses CLIP \cite{radford2021learning} text encoder to extract text embeddings from text prompts. Then, we take these text embeddings as the weight of a $1\times 1$ convolutional segmentation head to obtain the text-image logits. We use the text-image logits to enhance the original logits output to help the model better detect the anomaly classes in the image.}
\label{fig:mmras}
\end{figure*}

\section{Related Work}
\subsection{Anomaly segmentation}
In general, the anomaly segmentation approaches can be classified into three categories: (\romannumeral 1) \textbf{uncertainty-based}; (\romannumeral 2) \textbf{image resynthesis-based}; (\romannumeral 3) \textbf{outlier exposure-based}.

The \textit{uncertainty-based} approaches, originated from image-level Out-of-Distribution (OoD) detection, leverage statistics of logits and softmax probability to detect anomaly objects in the image. In \cite{hendrycks2017a}, Hendrycks et al. propose a strong baseline, named MSP, to detect image-level OoD data. Afterward, to scale MSP to large-scale datasets and real-world cases, they propose MaxLogits \cite{hendrycks2022scaling}, which even has applications in autonomous driving scenes (the CAOS benchmark, also proposed in \cite{hendrycks2022scaling}). To mitigate the degradation caused by the significant differences between the distribution of max logits of each predicted class, Jung et al. propose SML (Standardized Max Logits) \cite{jung2021standardized} to better apply MaxLogits \cite{hendrycks2022scaling} to segment anomalies in urban scenes. Besides, they also develop an iterative boundary suppression method to further improve the performance of SML. 

The \textit{image resynthesis-based} methods accomplish anomaly segmentation by using the reconstruction error between the input image and its reconstructions. The main challenge here is how to design an effective reconstruction framework. In \cite{creusot2015real}, Creusot et al. propose to detect the anomalies by using a Restricted Boltzmann Machine neural network to reconstruct the appearance of the road. In recent work, \cite{lis2019detecting} propose to reconstruct images from the semantic maps using generative networks \cite{wang2018high}. In SynBoost \cite{Di_Biase_2021_CVPR}, the authors propose an anomaly segmentation framework that produces robust predictions by combining resynthesis-based approaches with uncertainty-based approaches. 

Moreover, the \textit{outlier exposure-based} methods leverage auxiliary outlier datasets to train the network to improve its anomaly segmentation performance \cite{hendrycks2018deep}. In \cite{bevandić2019discriminative}, the authors propose to use outlier exposure to detect out-of-distribution pixels for semantic segmentation. In \cite{rai2023unmasking}, Rai et al. develop a mask contrastive learning approach to train the model using OoD data as supervision.

Based on the methods we discussed above, however, we find that no multi-modal-based anomaly segmentation solution for road scenes has been proposed so far. To the best of our knowledge, MMRAS+ is the first multi-modality uncertainty-based anomaly segmentation framework that innovatively introduces the text-modal to mitigate the issue of producing high anomaly scores for non-anomalous regions in the image.

\subsection{Mask-based anomaly segmentation}
In \cite{nayal2023rba}, Nayal et al. propose a scoring function, called Rba, which defines the event of being an unknown object as being rejected by all known objects. In \cite{grcic2023advantages}, the authors propose EAM for outlier-aware segmentation using mask-level recognition, achieving state-of-the-art performance by effectively leveraging mask-level uncertainty and incorporating negative data. In \cite{ackermann2023maskomaly}, the authors propose an anomaly segmentation framework by leveraging the raw mask outputs of the mask-based semantic segmentation networks. In \cite{rai2023unmasking}, they also study the application of mask-transformers \cite{cheng2022masked} in segmenting road anomalies. In their work, they improve the original Mask2Former \cite{cheng2022masked} architecture by developing a global attention module that allows the model to focus on both foreground and background information while maintaining the original inference speed.

Within our research, based on the mask-based architecture proposed in \cite{rai2023unmasking}, we take a further step to study the application of MaskCLIP \cite{zhou2022extract} in segmenting anomalies in autonomous driving scenes.

\section{Method}

\subsection{Overview}
As shown in Fig.~\ref{fig:mmras} and Fig.~\ref{fig:eb-module}, our framework mitigates the problem of producing high anomaly predictions for non-anomalous regions in the image by two key components: (\romannumeral 1) the Text Enhancement (T-E) module (Section \ref{sec:mmras}); (\romannumeral 2) the Ensemble Boosting (E-B) module (Section \ref{sec:mmras+}). In the T-E module, we extract text features using CLIP \cite{radford2021learning} text encoder and take the extracted text embeddings as the weight of the convolutional segmentation head to obtain the logits convoluted with text-modality. 

Then, we perform a weighted summation of the logits convoluted with text and the original logits output by the pre-trained model. In the E-B module, we ensemble different \textquotedblleft weak\textquotedblright~ anomaly scores to obtain a strong anomaly detector to boost the performance of segmenting anomalies. Next, we will introduce our multi-modal anomaly segmentation framework in detail with equations and figures.

\subsection{Multi-Modal Road Anomaly Segmentation}\label{sec:mmras}
Inspired by MaskCLIP \cite{zhou2022extract}, we propose to use CLIP \cite{radford2021learning} to segment the anomalies in the image. We follow the same procedure in \cite{zhou2022extract} to obtain text embeddings of classes in the Cityscapes \cite{cordts2016cityscapes} dataset (the inlier training data). To be specific, we first make 85 prompts using the class names of the in-distribution dataset. Then, we feed the prompts into the CLIP text encoder to obtain text embeddings. Afterward, we take these text embeddings as the weight of the convolution segmentation head to extract the logits convoluted with text (the text-image logits). Finally, we compute the weighted sum of the original logits value and the text-image logits to obtain the final logits for calculating the anomaly score. We name this multi-modal anomaly segmentation framework as MMRAS, the whole framework overview is shown in Fig.~\ref{fig:mmras}.

Given an input image $\mathbf{x}\in \mathbb{R}^{H\times W\times K}$, where $H$ is the height of the image, $W$ is the width of the input image, $K$ is the channel number. Denote the feature extracted by the image backbone as $\mathbf{f}$, the text embeddings extracted by CLIP text encoder as $\mathbf{t}\in \mathbb{R}^{C\times N}$, where $C$ is the number of classes, $N$ is the embedding dimension. Besides, let $\mathbf{m}$ be the logits value output by the Mask2Anomaly model. Then, the anomaly score calculated by MMRAS can be formulated as
\begin{align}
    \Tilde{\mathbf{f}} &= \phi_{\mathbf{t}} \left(\mathbf{f}\right),  \label{eq-1}\\
    \mathbf{l} &= \alpha \odot \mathbf{m}  + (1-\alpha) \odot \Tilde{\mathbf{f}},\quad \mathbf{l}\in \mathbb{R}^{C\times H\times W}, \label{eq-2}\\
    \textbf{MMRAS}\left(\mathbf{l}\right) &= 1-\max_{i=1}^C \mathbf{l}_i, \label{mmras-score}
\end{align}
where $\phi_{\mathbf{t}}$ is a $1\times 1$ convolutional layer taking $\mathbf{t}$ as its weight, $\Tilde{\mathbf{f}}$ is the text enhanced feature, $\alpha\geq 0$ is the summing weight to calculate the final logits output of the model, $\odot$ denotes the Hadamard product, $\mathbf{l}_i$ is the $i^{th}$ channel of logits $\mathbf{l}$. 

In addition, we dubbed equation (\ref{eq-1}) and (\ref{eq-2}) as the Text Enhancement (T-E) module, which we empirically showed its effectiveness in the experimental part of our paper.

\subsection{Boosting via Ensemble}\label{sec:mmras+}
To further improve the performance of segmenting anomalies, we develop a boosting method taking the weighted sum of MMRAS score and other anomaly scores \cite{hendrycks2022scaling}.

The main idea here is similar to AdaBoost \cite{freund1997decision}, which tries to obtain a strongly learnable algorithm by integrating several weakly learnable algorithms. Here, we want to create a strong anomaly segmentation score via ensembling many \textquotedblleft weak\textquotedblright~ anomaly scores. 

\begin{figure}[t]
\centering
\includegraphics[width=0.97\linewidth]{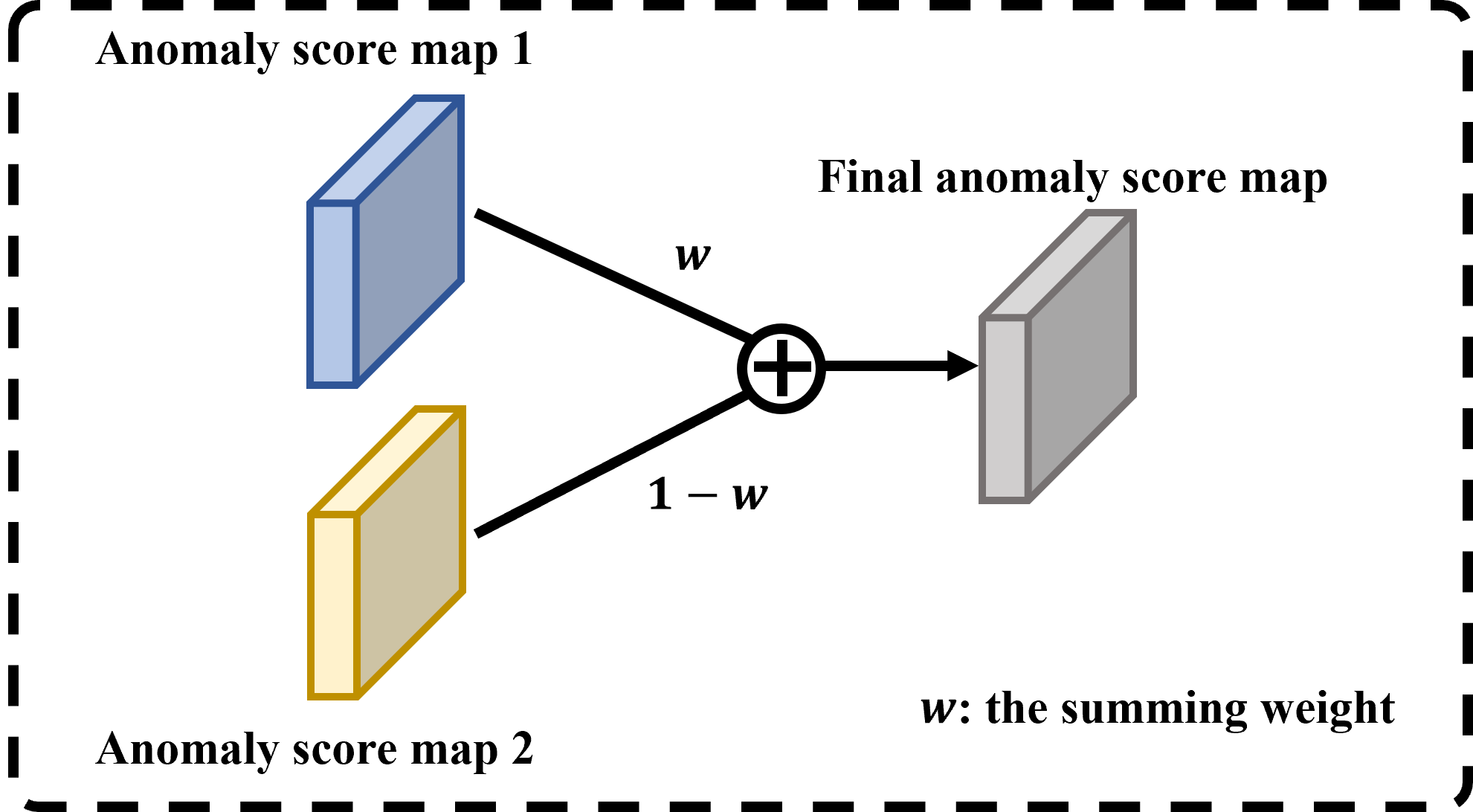}
\caption{The ensemble boosting module for enhancing the anomaly segmentation performance, inspired by AdaBoost \cite{freund1997decision}, which is essentially a weighted sum of different anomaly scores. By adding the E-B module to MMRAS, we obtain MMRAS+.}
\label{fig:eb-module}
\end{figure}

Specifically, based on the MMRAS score given in equation (\ref{mmras-score}), we further develop MMRAS+ as follows:
\begin{align}
    \textbf{MMRAS+}\left(\mathbf{l}, \mathbf{m}\right) &= w\odot\left(1-\max_{i=1}^C \mathbf{l}_i\right) \notag\\
    &+ (1-w)\odot\left(1-\max_{i=1}^C \mathbf{m}_i\right),\label{mmras+} 
\end{align}
where $w\geq 0$ is the summing weight. The intuitive interpretation for equation (\ref{mmras+}) is shown in Fig.~\ref{fig:eb-module}. We name equation (\ref{mmras+}) as the Ensemble Boosting (E-B) module. We also provide ablation analysis for this module in our experiments.

\begin{table*}[t]
\begin{center}
    \caption{Evaluation on RoadAnomaly \cite{lis2019detecting} and SMIYC \cite{chan2021segmentmeifyoucan} validation datasets. Herein, $\uparrow$ indicates the higher the value, the better the anomaly inference performance. The best results are in boldface.}
    \label{tab-validation}
    \resizebox{\textwidth}{!}{
    \begin{tabular}{ccccccccccccc}
    \toprule[2.5pt]
        \multirow{2}{*}{\textbf{Methods}} & \multicolumn{3}{c}{\textbf{RoadAnomaly}}  & \multicolumn{3}{c}{\textbf{SMIYC-RA21}} & \multicolumn{3}{c}{\textbf{SMIYC-RO21}} \\
          ~ & \textbf{FPR@95 $\downarrow$} & \textbf{AuPRC $\uparrow$} & \textbf{AuROC $\uparrow$} & \textbf{FPR@95 $\downarrow$} & \textbf{AuPRC $\uparrow$} & \textbf{AuROC $\uparrow$} & \textbf{FPR@95 $\downarrow$} & \textbf{AuPRC $\uparrow$} & \textbf{AuROC $\uparrow$} \\ 
          
        \midrule[1.5pt]
        MSP \cite{hendrycks2017a} & 14.48 & 67.91 & 94.97 & 3.30 & \textbf{93.55} & 98.94 & 0.75 & 57.36 & 99.37 \\
        Entropy \cite{hendrycks2017a} & 54.63 & 15.37 & 68.90 & 58.60 & 29.59 & 73.78 & 20.19 & 2.43 & 87.63 \\
        ODIN \cite{liang2018enhancing} & 32.86 & 55.03 & 91.37 & 3.58 & 91.54 & 98.45 & 1.05 & 47.79 & 99.15 \\
        MaxLogits \cite{hendrycks2022scaling} & 13.35 & 79.74 & 96.88 & 3.19 & 93.44 & 98.95 & 0.65 & 60.50 & 99.68 \\
        Mask2Anomaly \cite{rai2023unmasking} & 13.55 & 79.53 & 96.57 & 3.21 & 93.11 & 98.62 & 0.65 & \textbf{60.53} & \textbf{99.76}\\
        \midrule[1.5pt]
        MMRAS+ (Ours) & \textbf{13.35} & \textbf{82.65} & \textbf{96.94} & \textbf{3.19} & 93.44 &  \textbf{98.95} & \textbf{0.65} & 60.50 & 99.68 \\
        \bottomrule[2.5pt]
    \end{tabular}
    }
\end{center}
\end{table*}


\section{Experiments}
In this section, we first introduce the experiment setup, including the introduction of the anomaly inference datasets, the evaluation metrics, and the implementation details. Then, we compare our method with other outstanding baselines. Besides, we provide extensive ablation analysis of different modules in MMRAS+. 

\subsection{Setup}
\noindent\textbf{Datasets.}\quad
The model used for anomaly inference is pre-trained on the Cityscapes \cite{cordts2016cityscapes} dataset, which includes 2,975 training images, 500 validation images, and 1,525 for testing. Then, according to \cite{rai2023unmasking}, it is fine-tuned using outlier auxiliary datasets generated from MS-COCO \cite{lin2014microsoft}.

We evaluate the anomaly segmentation performance of our method on Road Anomaly \cite{lis2019detecting}, Segment Me If You Can (SMIYC) \cite{chan2021segmentmeifyoucan} and Fishyscapes \cite{blum2021fishyscapes} benchmarks. 

Concretely, the \textit{RoadAnomaly} validation dataset contains 60 images obtained from the Internet, including anomalous objects from different scales. 

The \textit{SMIYC} consists of two subsets: the RoadAnomaly21 (SMIYC-RA21) and the RoadObstacle21 (SMIYC-RO21). Specifically, the SMIYC-RA21 contains 10 images for validation and 100 images for testing. The SMIYC-RO21 includes 30 validation images and 327 test images. 

Moreover, the \textit{Fishyscapes} datasets include Fishyscape static (Static) and Fishyscapes Lost $\&$ Found (L$\&$F). The Static consists of 30 validation and 1000 test images, while the L$\&$F contains 100 validation and 275 test images. 

During the inference phase, we can only have access to the ground truth of the validation split in RoadAnomaly, RoadAnomaly21, RoadObstacle21, and Fishyscapes, but not the test split.

\begin{table*}[t]
\begin{center}
    \caption{Evaluation on Fishyscapes \cite{blum2021fishyscapes} validation datasets. Here, $\uparrow$ indicates the higher the value, the better the anomaly inference performance. The best results are in boldface.}
    \label{tab-fs-val}
    \resizebox{0.74\textwidth}{!}{
    \begin{tabular}{cccccccccc}
    \toprule[2.5pt]
        \multirow{2}{*}{\textbf{Methods}} & \multicolumn{3}{c}{\textbf{Static}}  & \multicolumn{3}{c}{\textbf{L$\&$F}}  \\
          ~ & \textbf{FPR@95 $\downarrow$} & \textbf{AuPRC $\uparrow$} & \textbf{AuROC $\uparrow$} & \textbf{FPR@95 $\downarrow$} & \textbf{AuPRC $\uparrow$} & \textbf{AuROC $\uparrow$} \\ 
          
        \midrule[1.5pt]
        MSP \cite{hendrycks2017a} & 64.78 & 84.93 & 90.42 & 82.57 & \textbf{1.35} &  86.14 \\
        Entropy \cite{hendrycks2017a} & 74.19  & 32.82 & 65.48 & 75.46 & 0.43 &  63.92 \\
        ODIN \cite{liang2018enhancing} & 62.42 & 82.32 & 89.45 & 81.10 & 1.43 & 86.28 \\
        MaxLogits \cite{hendrycks2022scaling} & 81.53 & 87.47 & 91.87 & 35.34 & 0.97 & 86.18 \\
        Mask2Anomaly \cite{rai2023unmasking} & 71.35 & 86.46 & 91.33 & \textbf{21.82} & 1.00 & 86.55 \\
        \midrule[1.5pt]
        MMRAS+ (Ours) & \textbf{58.48} & \textbf{87.53} & \textbf{92.33} & 25.11 & 1.21 & \textbf{90.38} \\
        \bottomrule[2.5pt]
    \end{tabular}
    }
\end{center}
\end{table*}

\noindent\textbf{Evaluation metrics.}\quad Following \cite{tian2022pixel,liu2023residual, rai2023unmasking}, we compute the Area Under Receiver Operating Characteristics (AuROC), the False Positive Rate at a true positive rate of 95$\%$ (FPR@95) and the Area under the Precision-Recall Curve (AuPRC) to evaluate the performance of our method.

\noindent\textbf{Baselines.}\quad We mainly compare our method with MSP \cite{hendrycks2017a}, Entropy \cite{hendrycks2017a}, ODIN \cite{liang2018enhancing}, MaxLogits \cite{hendrycks2022scaling} and Mask2Anomaly \cite{rai2023unmasking}. For ODIN \cite{liang2018enhancing}, the temperature is set to $T=3.0$ following the settings given in the SMIYC benchmark \cite{chan2021segmentmeifyoucan} code repository. Note that when conducting comparison experiments, none of the above methods require retraining or modifying the original model architecture, which ensures a fairer comparison.


\begin{figure}[t]
\centering
\includegraphics[width=\linewidth]{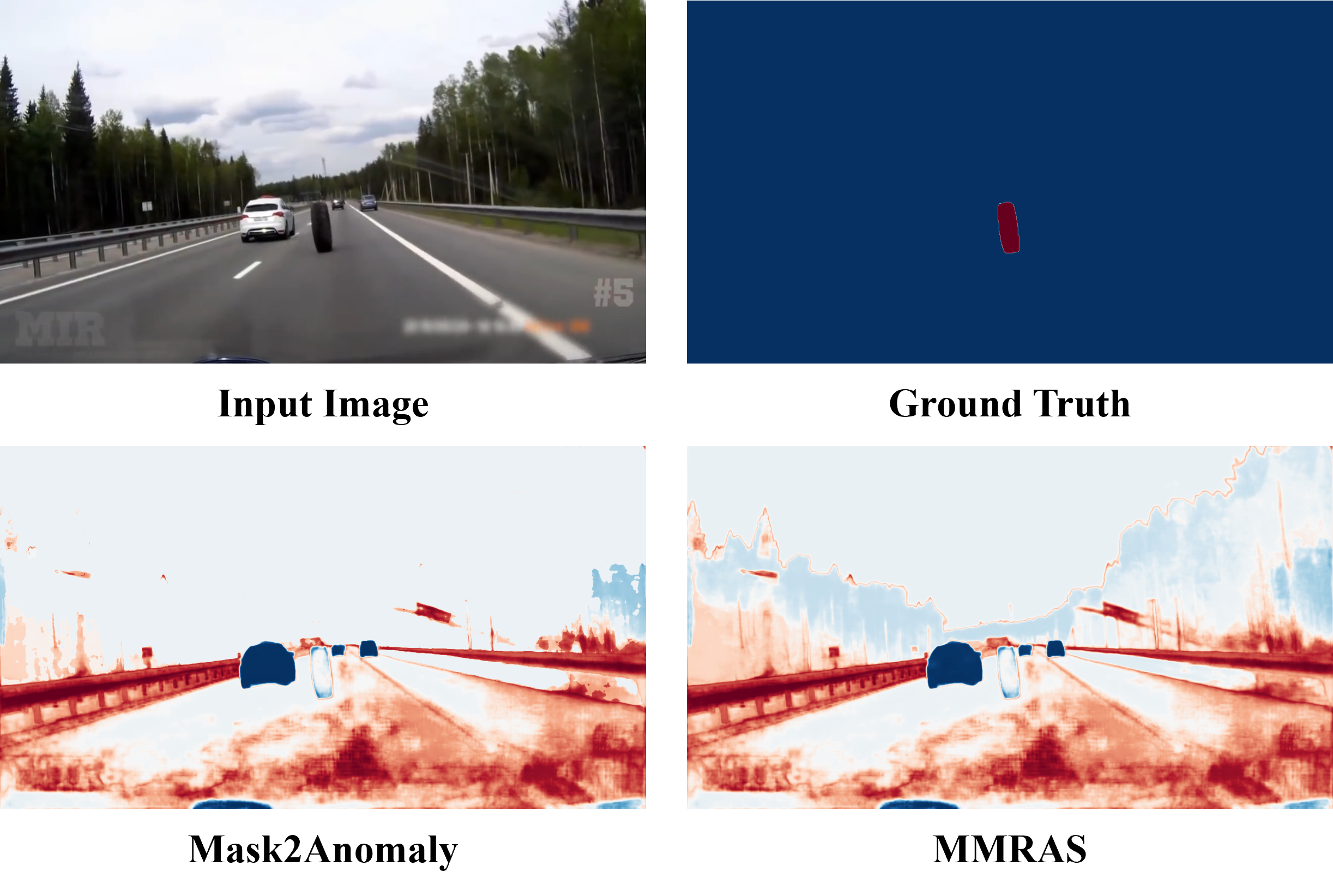}
\caption{The qualitative results of anomaly score maps on the RoadAnomaly validation dataset. By comparing the outputs of Mask2Anomaly and MMRAS, we can find that, after introducing the text-modal (MMRAS), the anomaly scores of the tree classes get markedly lower.}
\label{fig:qa-1}
\end{figure}

\noindent\textbf{Implementation details.}
To ensure fairness, all our experiments are conducted using a pre-trained Mask2Anomaly \cite{rai2023unmasking} model, which is developed based on a Mask2Former \cite{cheng2022masked} with ResNet-50 \cite{he2016deep} as the backbone and a decoder with global masked-attention. The pre-trained weight is given by the authors of Mask2Anomaly.
Note that different from the original work in \cite{rai2023unmasking}, \textbf{we do not use the anomaly score part in Mask2Anomaly \cite{rai2023unmasking}, but only its model part}. Besides, all methods are reproduced and evaluated on one NVIDIA V100 GPU with Python 3.7.16, CUDA 10.2 + Pytorch 1.10.0. 

The hyperparameter in equation (\ref{eq-2}) is set to $\alpha=0.99$ for RoadAnomaly \cite{lis2019detecting} validation dataset, $\alpha=0.9999999$ for SMIYC \cite{chan2021segmentmeifyoucan}, respectively. On the Fishyscape Static validation split, we set $\alpha=0.98$. Additionally, on Fishyscape Lost $\&$ Found validation split, we set $\alpha=0.7$. The reason why the hyperparameter $\alpha$ in equation (\ref{eq-2}) is close to 1 may be because the features obtained from the text embedding convolution can have a detrimental effect on the portion of the original image features that represent the anomalous object classes in the image. That is, the semantic maps segmented by CLIP may have some occlusion of the original objects in the feature map output by the original image encoder. Besides, the weight for ensemble boosting is set to $w=0.7$ for RoadAnomaly \cite{lis2019detecting}, SMIYC \cite{chan2021segmentmeifyoucan}, and Fishyscape Static validation datasets. On Fishyscape Lost $\&$ Found validation dataset, we set $w=0.9$.

\subsection{Evaluation results}
\noindent\textbf{Quantitative results.}\quad 
In this section, we provide the quantitative analysis for our method. We compare MMRAS+ with other approaches on RoadAnomaly \cite{lis2019detecting} and SMIYC \cite{chan2021segmentmeifyoucan} benchmarks. 

As shown in Table \ref{tab-validation}, based on Mask2Anomaly \cite{rai2023unmasking} architecture (only the model part), MMRAS+ outperforms other competitive methods under the same setting. Note that our method surpasses the SOTA method Mask2Anomaly \cite{rai2023unmasking} on RoadAnomaly \cite{lis2019detecting} by 3.12$\%$ on AuPRC, 0.2$\%$ on FPR@95 and 0.37$\%$ on AuROC, respectively. Meanwhile, MMRAS+ outperforms MSP \cite{hendrycks2017a} on RoadAnomaly validation split by 14.74$\%$ on AuPRC, 1.13 $\%$ on FPR@95 and 1.97$\%$ on AuROC.

In Table \ref{tab-fs-val}, we also find that MMRAS+ outperforms other competitive methods with a certain margin. For instance, on the Fishyscape Static validation dataset, MMRAS+ outperforms Mask2Anomaly by 12.87$\%$ in  FPR@95 and 1.00$\%$ in AuROC. On Fishyscape Lost $\&$ Found validation dataset, our method surpasses Mask2Anomaly by 3.83$\%$ in AuROC.

\begin{table}[t] 
\begin{center}
\caption{Ablation study of different modules in MMRAS+ on RoadAnomaly validation set. \lq T-E\rq~ denotes Text Enhancement, \lq E-B\rq~denotes Ensemble Boosting.} 
\label{tab-ablation}
    \resizebox{0.49\textwidth}{!}{
\begin{tabular}{c|cc|c|c|c}
  \toprule[2.5pt]
    & \makecell[c]{T-E} & \makecell[c]{E-B} & \textbf{FPR@95}$\downarrow$ & \textbf{AuPRC} $\uparrow$ & \textbf{AuROC} $\uparrow$
  \\
  \midrule[1.5pt]
  (\romannumeral 1)&  \XSolidBrush & \XSolidBrush & 13.35 & 79.74 & 96.88 \\
  (\romannumeral 2)&  \Checkmark & \XSolidBrush  &  13.35  & 82.48 & 96.94 \\
  (\romannumeral 3)&  \XSolidBrush & \Checkmark & 13.35 & 79.74 & 96.88 \\
  (\romannumeral 4)&  \Checkmark & \Checkmark & \textbf{13.35} & \textbf{82.65} & \textbf{96.94} \\
  \bottomrule[2.5pt]
\end{tabular}
}
\end{center}
\end{table}

\noindent\textbf{Qualitative results.}\quad
To verify the effectiveness of using text modal to prevent the model from producing high anomaly scores for the non-anomalous regions in the image. We visualize the anomaly score maps of MMRAS (without using the ensemble boosting technique) and Mask2Anomaly, comparing them with the Ground Truth in Fig. \ref{fig:qa-1}. From Fig. \ref{fig:qa-1}, we can observe that after introducing text modal into the original unimodal anomaly segmentation (the Mask2Anomaly), the anomaly scores for the trees in the image become much lower (the color is much closer to blue).

\subsection{Ablation study}
\noindent\textbf{Quantitative results.}\quad To analyze the impact of each module in MMRAS+, we conduct a complete ablation analysis for our multi-modal anomaly segmentation framework. 

As shown in Tab.\ref{tab-ablation}, setting (\romannumeral 2) proves the enhancement ability of our T-E module. Based on setting (\romannumeral 2), we add the E-B module (setting (\romannumeral 4)), which has a limited improvement compared with setting (\romannumeral 2).

\begin{figure}[t]
\centering
\includegraphics[width=\linewidth]{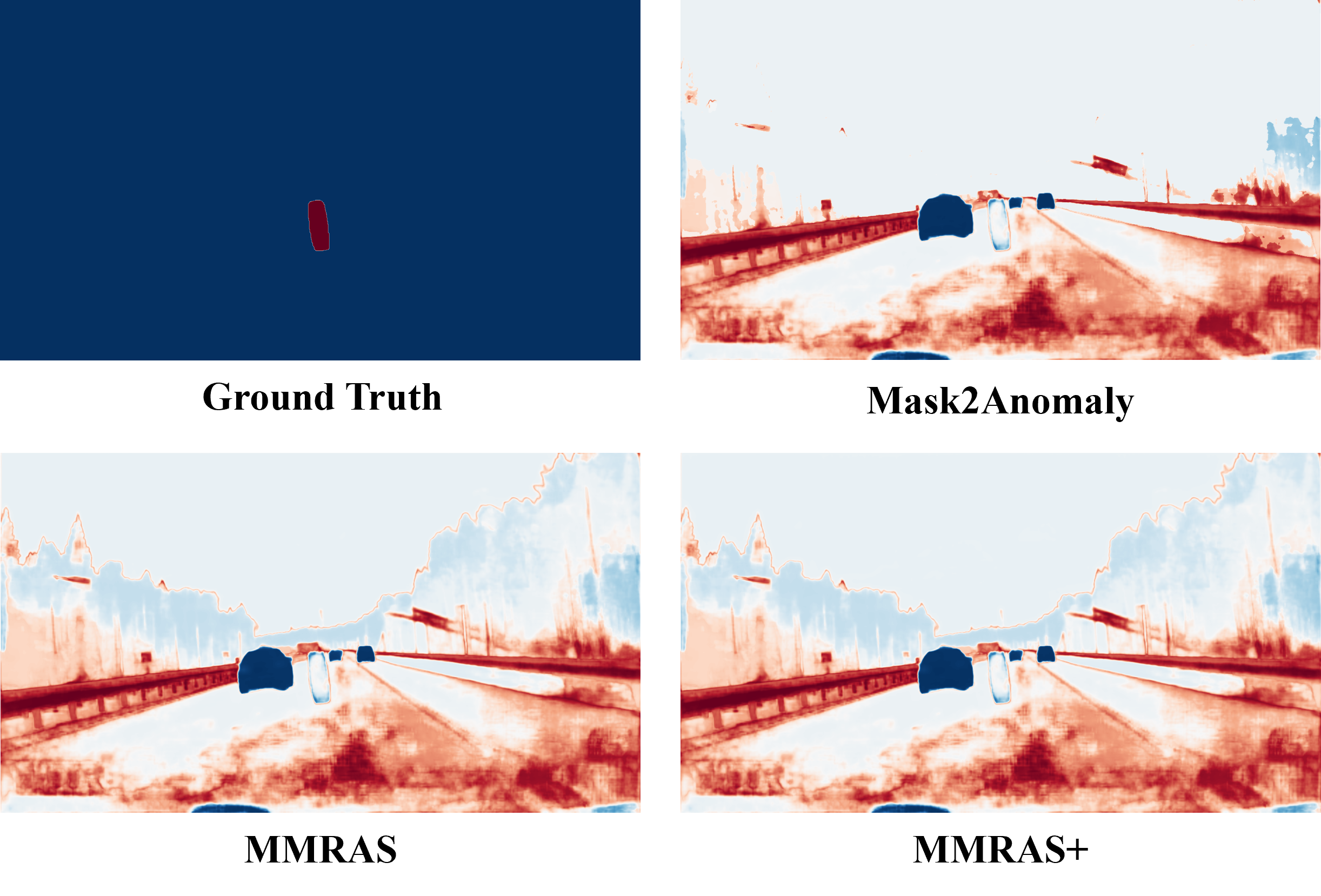}
\caption{Qualitative ablation analysis for MMRAS+ on RoadAnomaly \cite{lis2019detecting} validation dataset.}
\label{fig:mmras-ablation}
\end{figure}

\noindent\textbf{Qualitative results.}\quad We further visualize the anomaly outputs of MMRAS and MMRAS+ in Fig.\ref{fig:mmras-ablation}, comparing with the original Mask2Anomaly's outputs and the Ground Truth. From Fig.\ref{fig:mmras-ablation}, we can find that the improvement of MMRAS+ is not very significant. We can only observe the enhancement effect of ensemble boosting from the data presented in Tab.\ref{tab-ablation}.



\section{Conclusion}
In this paper, we presented MMRAS+, the first multi-modal uncertainty-based anomaly segmentation framework for autonomous driving scenes. We mitigate the issue of producing high anomaly scores for non-anomalous regions in driving scenes by introducing text-modal into the original uni-modal-based anomaly segmentation framework for autonomous driving systems. We studied how to leverage CLIP to segment anomalies in road scenes. Moreover, we develop an ensemble module to further boost the performance. The qualitative and quantitative experiments on RoadAnomaly, SMIYC, and Fishyscapes validation datasets demonstrate the superior performance of our method.

\bibliographystyle{IEEEbib}
\bibliography{icme2025references}


\end{document}